\documentclass[letterpaper,10pt,twoside, %
  electronic,
  papers=countpages, 
  paperselec=all, 
  hyperref={bookmarksdepth=1,bookmarksopen,bookmarksopenlevel=0,%
    colorlinks=false},
  geometry={text={175truemm,226truemm},
    inner=0.805in,top=29.15mm,bottom=24.5mm,footskip=9.68mm,voffset=-5mm},
]{confproc}
\usepackage[utf8]{inputenc}
\usepackage[T1]{fontenc}
\usepackage{mathptmx}
\usepackage[super]{nth}

\usepackage{longtable}

\renewcommand{\procpdfauthor}{{Sarah Alice Gaggl,  Juan Carlos Nieves, Hannes Strass (Eds.)}}
\renewcommand{\procpdftitle}{{Arg-LPNMR 2016 Proceedings}}

\author{\procpdfauthor}
\title{\procpdftitle}
\date{November 2016}

\makeindex

\begin{document}
\frontmatter
\setcounter{page}{1}
\pdfbookmark[0]{Preamble}{preamble}
\pdfbookmark[1]{Cover}{cover}
\maketitle

\section*{Preface}
This volume contains the papers presented at Arg-LPNMR 2016: First International Workshop on Argumentation in Logic Programming and Nonmonotonic Reasoning held on July 8-10, 2016 in New York City, NY.

Research on argumentation and an Artificial Intelligence (AI) began in full force in the early eighties. The initial efforts showed how argumentation results in a very natural way of conceptualizing commonsense reasoning, appropriately reflecting its defeasible nature. In the mid-nineties, Dung (1995) has shown that argumentation provides a useful perspective for relating different non-monotonic formalisms. Currently, argumentation has been applied in different subfields of AI like Multi-Agent Systems, Semantic Web, knowledge representation and reasoning, etc.

Works in the knowledge representation and reasoning community have shown that argumentation inferences in terms of the so called argumentation semantics have strong roots in logic-based theories and non-monotonic reasoning. In this sense, the relationship between logic programming and argumentation has attracted increased attention in the last years. Studies range from translating one into the other and back, using argumentation to explain logic programming models, and using logic programming systems to implement argumentation-based languages (ASPARTIX, DIAMOND). Influences go both ways and we believe that both fields can benefit from learning from each other. Moreover, argumentation allows to relate several non-monotonic formalisms such as belief revision, reasoning about actions and probabilistic reasoning.

More recently, argumentation has been revealed as a powerful conceptual tool for exploring the theoretical foundations of reasoning and interaction in autonomous systems and Multi-Agent Systems. Different dialogue models have been proposed based on the roots of argumentation. Indeed considering argumentation roots, the so called Agreement Technologies have been suggested in order to deal with the new requirement of interaction between autonomous systems and Multi-Agent Systems.

The workshop centered around four current research strands in abstract argumentation, namely analyzing argumentation semantics, studying dynamics in argumentation, and implementations of systems for abstract argumentation.

The committee decided to accept 2 papers. The program also includes 3 invited talks.

~\bigskip

\noindent
\begin{minipage}[t]{.4\textwidth}
November 2016
\end{minipage}%
\hfill
\begin{minipage}[t]{.4\textwidth}\flushright
Sarah Alice Gaggl\\
Juan Carlos Nieves\\
Hannes Strass
\end{minipage}

\newpage

\section*{Program Committee}
\noindent
\begin{longtable}{p{0.35\textwidth}p{0.65\textwidth}}
Gerhard Brewka & Leipzig University\\
Sylvie Doutre & University of Toulouse 1 - IRIT\\
Dov Gabbay & King's College, London\\
Sarah Alice Gaggl & Technische Universit\"at Dresden\\
Gabriele Kern-Isberner & Technische Universitaet Dortmund\\
Juan Carlos Nieves & Ume\r{a} University\\
Mauricio Osorio & UDLAP\\
Phan Phan Minh Dung & AIT\\
Chiaki Sakama & Wakayama University\\
Jan Sefranek & Comenius University, Bratislava\\
Guillermo Simari & Dep of Computer Science and Engineering, Universidad Nacional del Sur in Bahia Blanca\\
Hannes Strass & Leipzig University\\
Paolo Torroni & University of Bologna\\
Toshiko Wakaki & Shibaura Institute of Technology\\
Stefan Woltran & TU Wien\\
\end{longtable}

\otherpagestyle
\tableofcontents

\mainmatter

  \session{Invited Talks}
    \procpaper[switch=45,  npages=3,%
      title={SAT-Based Approaches to Reasoning about Argumentation Frameworks},%
      author={Matti J\"arvisalo},%
      index={\index{J\"arvisalo, Matti}},%
    ]{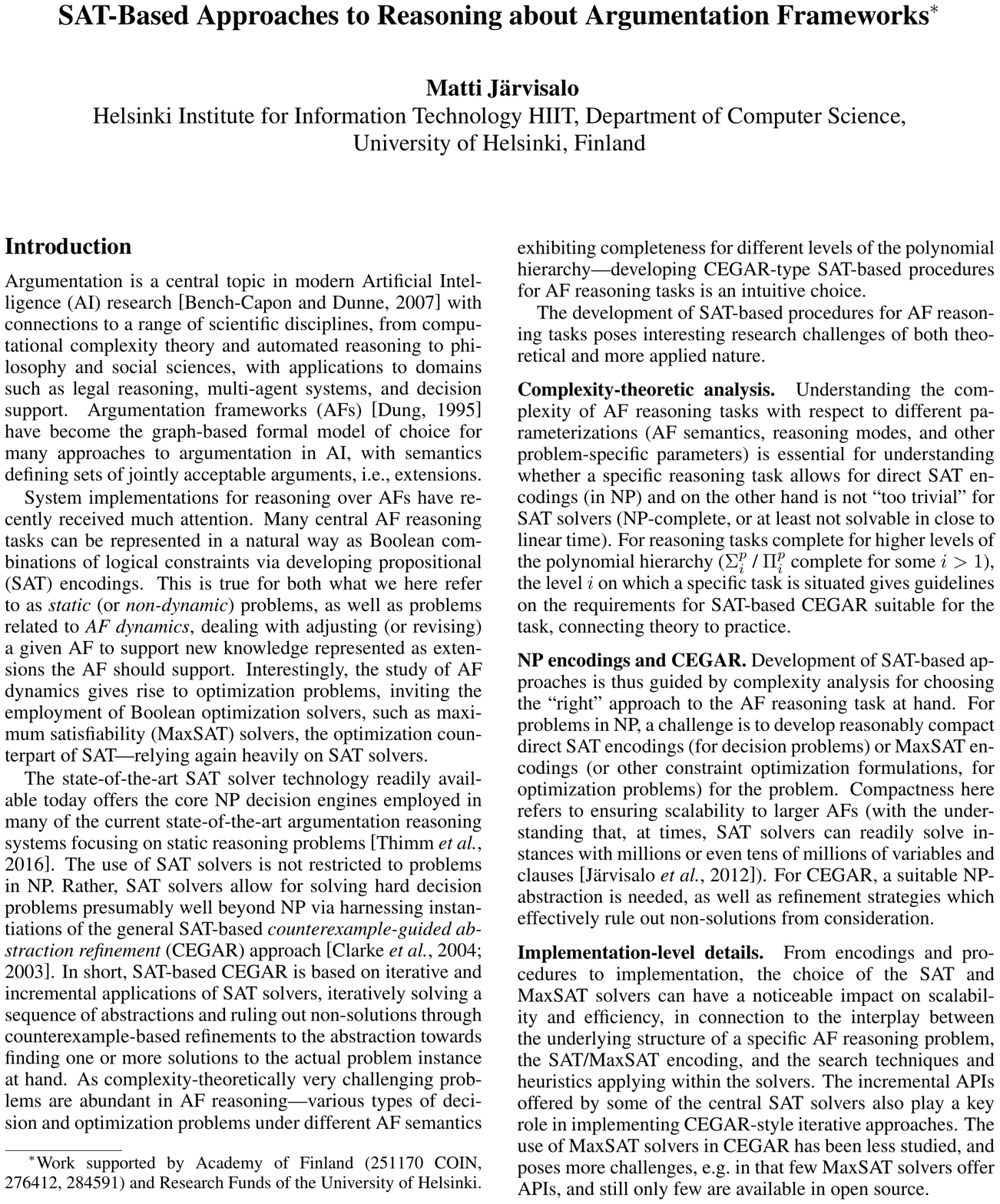}
    \procpaper[switch=21,  npages=2,%
      title={Acceptability semantics for argumentation frameworks},%
      author={Leila Amgoud},%
      index={\index{Amgoud, Leila}},%
    ]{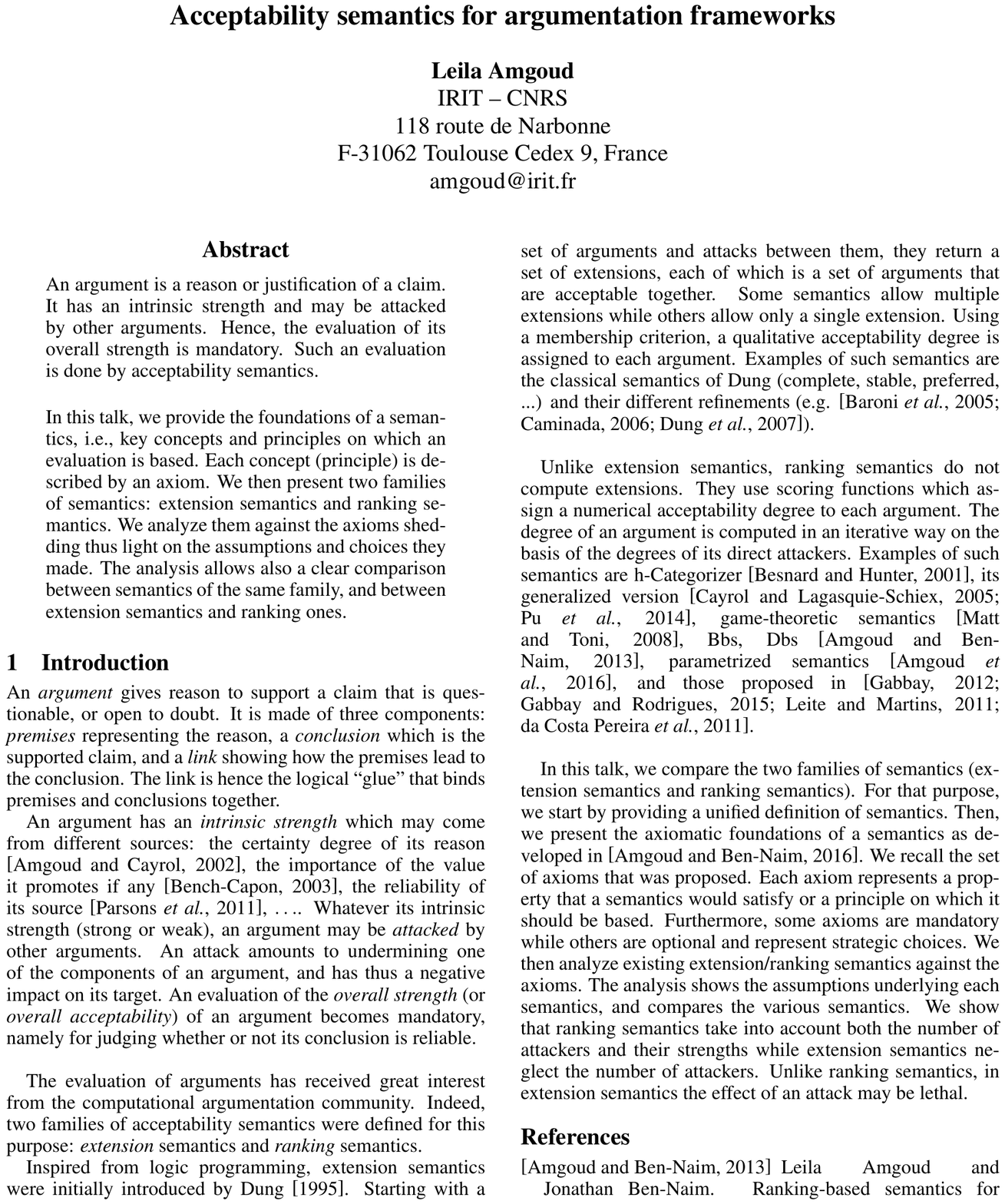}
    \procpaper[switch=33,%
      title = {Aggregating Opinions in Abstract Argumentation},%
      author={Richard Booth},%
      index={\index{Booth, Richard}},%
    ]{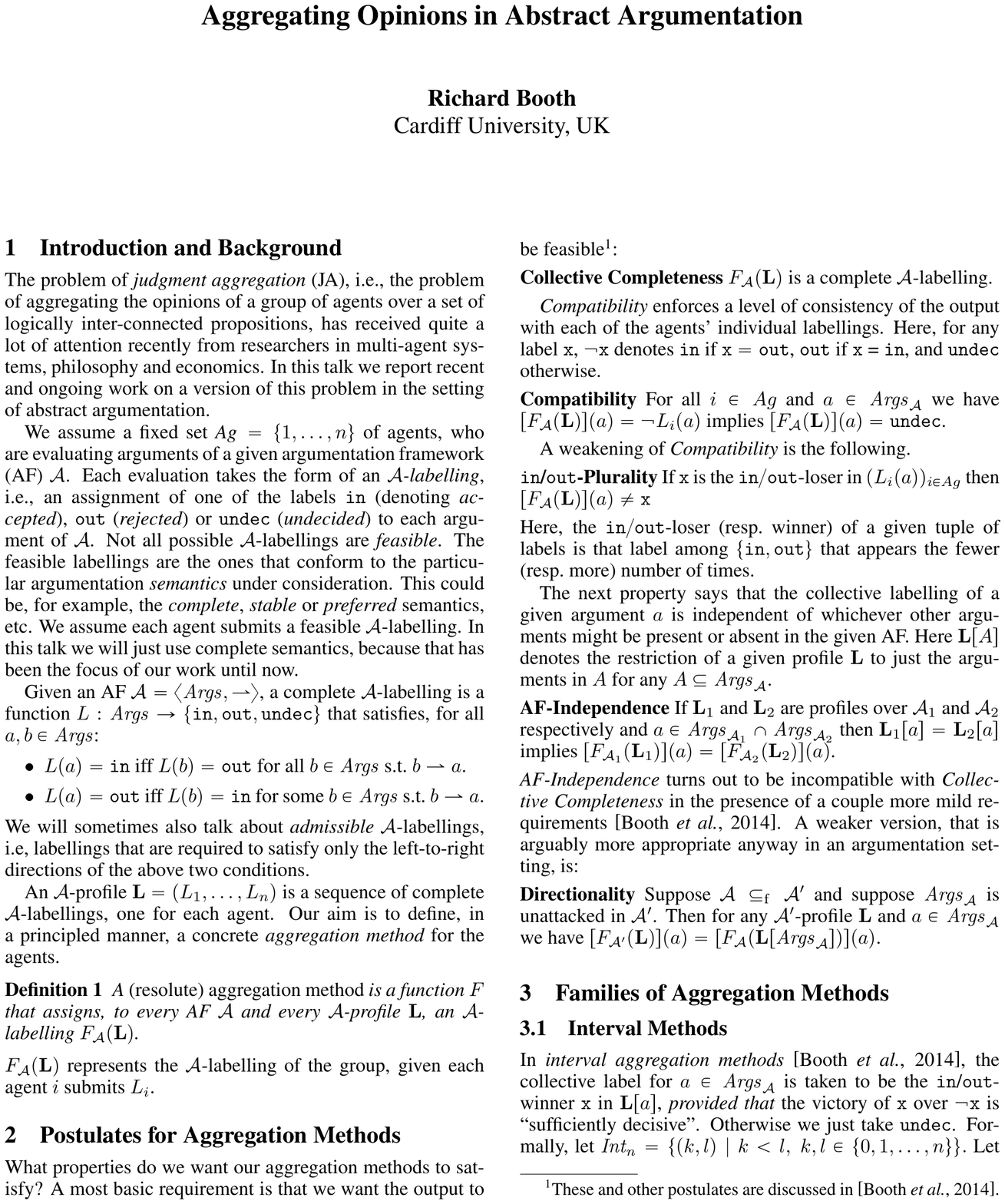}

  \session{Technical Talks}
    \procpaper[switch=75,%
      title={Collaborative Planning and Decision Support for Practical Reasoning in Decentralized Supply Chains},%
      author={Naeem Janjua},%
      index={\index{Janjua, Naeem}},%
    ]{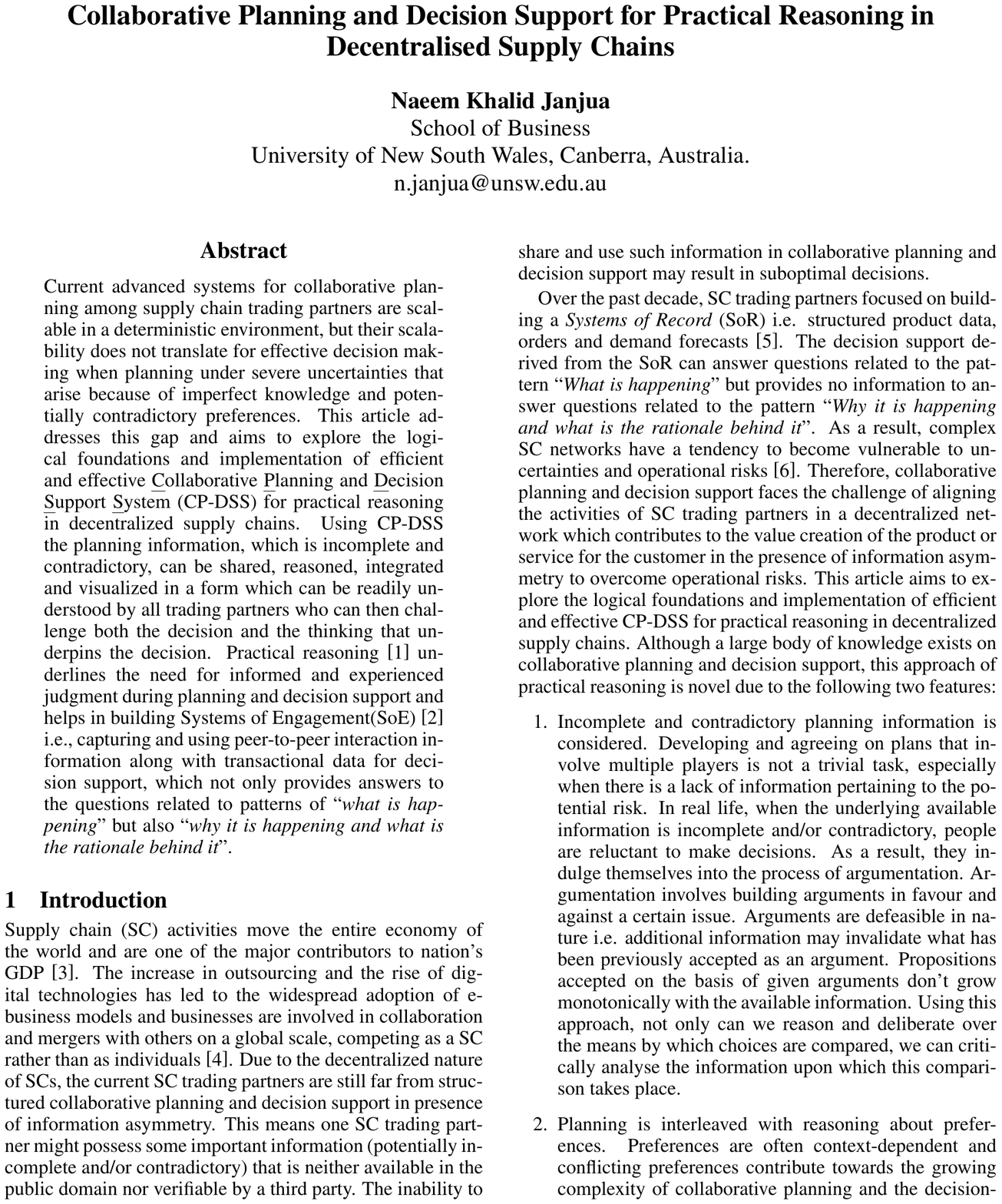}
    \procpaper[switch=27,%
      title={Revision of Abstract Dialectical Frameworks: Preliminary Report},%
      author={Thomas Linsbichler, Stefan Woltran},
      index={\index{Linsbichler, Thomas}\index{Woltran, Stefan}},
    ]{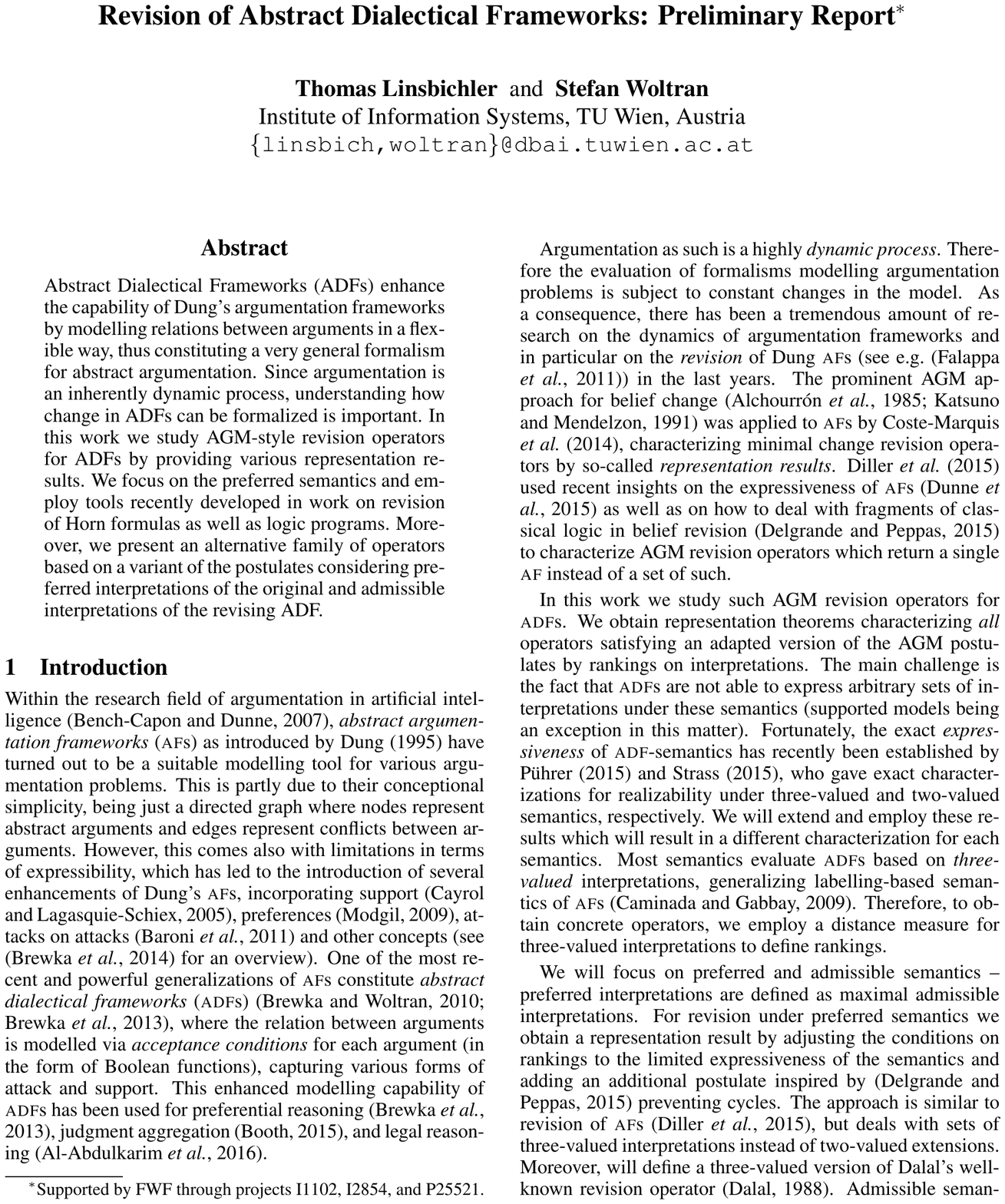}

\backmatter
\end{document}